\documentclass[10pt, a4paper]{article}

\usepackage{lrec-coling2024} 

\usepackage{ulem}
\usepackage{kotex}
\usepackage{float}

\usepackage{amsmath}
\usepackage{amsfonts}
\usepackage{mathrsfs}
\usepackage{comment}
\usepackage{makecell}

\usepackage{tabularx}
\usepackage{multirow}
\usepackage{boldline}
\usepackage{array}
\usepackage{stackengine}
\usepackage{arydshln}
\usepackage{amsmath}
\usepackage{graphicx}

\usepackage{natbib}
\usepackage{multibib}
\makeatletter
\def\@mb@citenamelist{cite,citep,citet,citealp,citealt,citepalias,citetalias}
\makeatother
\newcites{languageresource}{~}

\usepackage{graphicx}
\usepackage{tabularx}
\usepackage{soul}
\usepackage{titlesec}
\titleformat{\section}{\normalfont\large\bfseries\center}{\thesection.}{1em}{}
\titleformat{\subsection}{\normalfont\bfseries\raggedright}{\thesubsection.}{1em}{}
\titleformat{\subsubsection}{\normalfont\normalsize\bfseries\raggedright}{\thesubsubsection.}{1em}{}
\renewcommand\thesection{\arabic{section}}
\renewcommand\thesubsection{\thesection.\arabic{subsection}}
\renewcommand\thesubsubsection{\thesubsection.\arabic{subsubsection}}

\usepackage{xcolor}
\usepackage{hyperref}
 \definecolor{darkblue}{rgb}{0, 0, 0.5}
  \hypersetup{colorlinks=true, citecolor=darkblue, linkcolor=darkblue, urlcolor=darkblue}

\usepackage{xstring}

\usepackage{color}

\title{Optimizing Language Augmentation for Multilingual Large Language Models: A Case Study on Korean}
\newcommand{\authorname}[1]{{\fontsize{12pt}{14pt}\selectfont\textbf{#1}}}

\name{ChangSu Choi$^{1}$$^{\ddagger}$\thanks{$^{\ddagger}$These authors contributed equally.}, 
Yongbin Jeong$^{3}$$^{\ddagger}$, 
Seoyoon Park$^{2}$$^{\ddagger}$, \\
\authorname{InHo Won$^{1}$,
HyeonSeok Lim$^{1}$,
SangMin Kim$^{1}$,
Yejee Kang$^{3}$},\\
\authorname{Chanhyuk Yoon$^{3}$,
Jaewan Park$^{3}$,
Yiseul Lee$^{3}$, HyeJin Lee$^{4}$, }\\
\authorname{Younggyun Hahm$^{3}$, Hansaem Kim$^{2}$ and KyungTae Lim$^{1}$$^{\dagger}$}\thanks{$^{\dagger}$Corresponding author.}}

\address{$^{1}$SeoulTech, $^{2}$Yonsei University, $^{3}$Teddysum, $^{4}$KISTI \\
         choics2623@seoultech.ac.kr, ybjeong@teddysum.ai, seoyoon.park@yonsei.ac.kr\\
         \{wih1226, gustjrantk, sangmin6600, ktlim\}@seoultech.ac.kr\\
         khss@yonsei.ac.kr,
         \{kangyj, chyoon, jwpark, yslee, hahmyg\}@teddysum.ai,
         hyejin@kisti.re.kr}

\abstract{
Large language models (LLMs) use pretraining to predict the subsequent word; however, their expansion requires significant computing resources. Numerous big tech companies and research institutes have developed multilingual LLMs (MLLMs) to meet current demands, overlooking less-resourced languages (LRLs). This study proposed three strategies to enhance the performance of LRLs based on the publicly available MLLMs. First, the MLLM vocabularies of LRLs were expanded to enhance expressiveness. Second, bilingual data were used for pretraining to align the high- and less-resourced languages. Third, a high-quality small-scale instruction dataset was constructed and instruction-tuning was performed to augment the LRL. The experiments employed the Llama2 model and Korean was used as the LRL, which was quantitatively evaluated against other developed LLMs across eight tasks. Furthermore, a qualitative assessment was performed based on human evaluation and GPT4. Experimental results showed that our proposed \texttt{Bllossom} model exhibited superior performance in qualitative analyses compared to previously proposed Korean monolingual models.
 \\ \newline \Keywords{Large Language Model, Less-resourced Languages, Instruction-tuning} }

\begin{document}

\maketitleabstract

\section{Introduction} \label{sec:intro}
A large language model (LLM) comprehends linguistic information and knowledge via pretraining to predict the subsequent word based on the given context~\cite{zhao2023survey}.
However, the growth of LLMs increases the computing resources required for training, posing a challenge for smaller research groups to utilize them realistically~\cite{NEURIPS2022chinchilla}. To meet the demands of this era, numerous big tech companies and research institutes have been competing to launch multilingual LLMs~(MLLMs)~\cite{touvron2023Llama1,touvron2023Llama2,scao2022bloom}. However, less-resourced languages (LRLs) are being overlooked~\cite{gu-etal-2018-universal}. 
The recently launched Llama2~\cite{touvron2023Llama2} is an MLLM trained in more than 28 languages; however, only 0.06\% of the data was used for the Korean language. This leads to  two significant syntactic and semantic challenges. First, during the MLLM training, LRLs use minimal vocabulary based on the scarce training data, which limits their expression owing to the inadequate lexicon. Second, greater semantic knowledge is required to employ LLMs for specific tasks, such as question-answering, thereby rendering the models inapplicable or prone to hallucinations~\cite{zheng2023does,peng2023check}.

Numerous methods have been proposed to enhance the LRL performance. These include expanding the vocabulary of word embeddings~\cite{wang2019improving,schuster2019crosslingual}, aligning multilingual embeddings by combining them with other languages~\cite{artetxe-etal-2017,artetxe-etal-2018}, and reinforcing the utility of LLMs using minimal training data. The \textbf{L}ess \textbf{I}s \textbf{M}ore for \textbf{A}lignment~(LIMA) study proposed a method to maximize the utility of LLMs using 1,030 high-quality instruction data\cite{zhou2023lima}. 

Based on the existing research, it remains to be investigated whether MLLMs can expand their vocabulary and enhance the semantic inferencing capability of a specific language using minimum additional data.

This study explores the aforementioned aspects by proposing a method to enhance the Korean language capabilities of a representative MLLM, i.e., Llama2. The specific language abilities of the MLLM are enhanced using the following strategies: 
(1) \textbf{Vocabulary expansion}: To enhance its fundamental vocabulary, the MLLM was augmented with the dictionary of a specific language.
(2) \textbf{Knowledge enrichment}: The vocabulary and knowledge information of this language were enhanced in the MLLM via pretraining.  
(3) \textbf{Usability enhancement}: High-quality instruction data were generated in the Korean language to improve its LLM applicability. Korean is not an LRL as it comprises various language resources and evaluation data~\cite{park2021klue}. However, Korean is experimentally suitable as a relatively less-resourced language because the Llama2 model uses limited Korean data and vocabulary during training.

To enhance the vocabulary, knowledge reinforcement, and usability, 7,478 Korean vocabulary entries were added and pretraining was performed using a Korean–English corpus. The 1,030 English data, proposed by LIMA ~\cite{zhou2023lima}, were restructured by three Korean linguistics to ensure their practical similarity with the Korean language, thus enhancing their usability. 

The effectiveness of the three enhancement methods were validated by addressing the following questions: (1) What are the advantages of expanding the Korean vocabulary? (2) Is it effective to connect the knowledge of high- and low-resource languages via pretraining? (3) Do the actual usability and accuracy improve using the proposed Korean LIMA data? We propose the \texttt{Bllossom} model that applies the three aforementioned methods. Quantitatively, this model demonstrated an average performance improvement from 1.8\% to 8\% across eight tasks compared to the model without vocabulary expansion. For the qualitative evaluation of each model, the answers to 300 queries were compared using human and preference evaluations based on GPT4. Consequently, the \texttt{Bllossom} model outperformed the other Korean models of the same size by 93\%. The contributions of this study are as follows:

\begin{itemize}
    \item{A method for enhancing the vocabulary, knowledge, and usability of LRLs using MLLMs was proposed.}
    \item{A method for constructing instruction data based on language-specific features was presented and demonstrated by constructing a Korean LIMA dataset.}
    \item{For easy utilization, the data, models, and services used to construct and evaluate the Korean LLM were made publicly accessible\footnote{\href{https://github.com/Anonymous-introduce-paper/LR}{github.com/Anonymous-introduce-paper/LR}}.}
\end{itemize}

\section{Related Work}
\label{sec:related_work}
\subsection{MLLMs}
LLMs are massive pretrained language models containing more than several hundred billions of parameters ~\cite{zhao2023survey}. The generative LLMs based on the decoders of various transformers ~\cite{vaswani2017transformer}, including GPT series models~\cite{radford2018gpt-1, radford2019gpt-2, brown2020gpt-3}, are pretrained using the causal language modeling (CLM) approach. This approach predicts the subsequent token based on the preceding token sequence, providing insights into language, grammar, and knowledge. However, the user intent is difficult to determine because CLM can only predict the subsequent token. Hence, a pretrained model should be tuned to accurately comprehend the user intent and achieve capabilities, such as instruction adherence~\cite{NEURIPS2022_InstructGPT}.

This tuning process is referred to as “instruction tuning”, which is commonly implemented using supervised fine tuning (SFT) that considers the two directives and input data as model inputs and predicts the output data~\cite{wei2022flan}. SFT maximizes the training efficiency using a small amount of high-quality data rather than a large amount of low-quality data. LIMA~\cite{zhou2023lima} uses a smaller amount of highly refined SFT training data of 1,030 to outperform models trained on large amounts of auto-generated or low-quality SFT data. Additionally, LIMA proposes qualitative performance evaluation methods including, human and GPT4 evaluation for LLMs to determine the superior model.

Multilingual LLMs are advantageous in accumulating vast training data from multiple languages. Owing to the increasing parameters and LLM training data, models such as FLAN-T5~\cite{wei2022flan}, BLOOM~\cite{scao2022bloom}, Falcon, Llama~\cite{touvron2023Llama1}, and Llama2~\cite{touvron2023Llama2}, which are smaller but comprehensively acquire the multilingual knowledge, have attracted scholarly attention. Llama2 is a multilingual language model trained using large-scale publicly available data (including CommonCrawl, Github, Wikipedia, and ArXiv) for more than 28 languages. Thus, it possesses cross-language understanding capabilities. However, languages with non-Latin character systems, such as Korean and Chinese, exhibited inferior performances~\cite{cui2023efficient}. 

\subsection{Open-source Korean LLMs}

EleutherAI developed Polyglot-Ko, which is a monolingual Korean LLM pretrained on 1.2 TB of Korean data and contains models with sizes up to 12.8B~\cite{ko2023polyglot-ko}. KoAlpaca\footnote{\href{https://github.com/Beomi/KoAlpaca}{https://github.com/Beomi/KoAlpaca}} is a model based on Polyglot-Ko that automatically translates the SFT data of the English Alpaca~\cite{taori2023alpaca} into Korean and performs SFT with a total of 21K data. Similarly, Kullm~\cite{2023kullm} was proposed based on Polyglot-Ko and tuned using additional instruction data. Kullm used 153K SFT training data by translating English SFT datasets, including the GPT-4-LLM~\cite{peng2023instruction}, Vicuna~\cite{vicuna2023}, and Dolly from Databricks~\cite{DatabricksBlog2023DollyV2}. Additionally, models that perform the Korean SFT based on multilingual models, such as Llama2, have been launched. Komt is an instruction-tuned model based on Llama2 using a total of 1,543K data processed from existing Korean natural language processing data. Ko-Platypus2~\cite{lee2023platypus} enhances the logic knowledge of LLMs using a translated dataset from English Open-Platypus into Korean. This model is tuned using Llama2 with 25K SFT data. The aforementioned Korean models were used in the experiment, and the access links and summary information for each model are listed in Table~\ref{tb-KoreanLLMs}.

\section{Enriching the MLLM vocabulary}
This section introduces the following two approaches to the three language enhancement methods proposed in the Introduction: (1) vocabulary expansion and (2) knowledge enrichment. We propose a method to expand the Korean vocabulary in Llama2, which is a representative multilingual LLM, and reinforce the knowledge information between the Korean and English languages via CLM-based pretraining.

\begin{table}
\small
\centering
\begin{tabular}{l|p{5.6cm}} 
\hlineB{4}
\multicolumn{2}{c}{Sentence: 햄버거를 먹는 공룡} \\
\multicolumn{2}{c}{(A dinosaur eating a hamburger)} \\
\Xhline{1pt}
\textbf{Model} & \multicolumn{1}{c}{\textbf{Tokenization results}} \\
\Xhline{1pt}
Llama2 & `\textunderscore', `<0xED>', `<0x96>', `<0x84>', `<0xEB>', `<0xB2>', `<0x84>', `<0xEA>', `<0xB1>', `<0xB0>', `를', `\textunderscore', `<0xEB>', `<0xA8>', `<0xB9>', `는', `\textunderscore', `공', `<0xEB>', `<0xA3>', `<0xA1>' \\
\hline
Proposed & `햄', `버', `거', `를', `\textunderscore먹는', `\textunderscore', `공', `룡' \\
\hlineB{4}
\end{tabular}
\caption{\label{tb-tokenizing-result} Comparison of tokenization results between Llama2 and the proposed model}
\end{table}

\subsection{Vocabulary expansion}

The training data of Llama2 consisted of 89.7\% English words, and the tokenizer dictionary~($\mathcal{D}_{L}$) was composed of 90\% English (or Latin) words. The majority of the remaining words were rare words, neologisms, and LRLs categorized as out-of-vocabulary (OOV). To address this issue, Llama2 employed the SentencePiece tokenizer~\cite{kudo2018sentencepiece} that uses a UTF-8 byte fallback mechanism to handle the OOV words by decomposing them to the UTF-8 byte level. Therefore, the words not in $\mathcal{D}_{L}$ are represented without expanding the tokenizer vocabulary. 
The Korean vocabulary was expanded despite having a method to represent the language. Table~\ref{tb-tokenizing-result} compares the tokenizing results of Llama2 and the proposed model with an expanded Korean vocabulary for the sentence “햄버거를 먹는 공룡”. In the original Llama2, the Korean word “햄” was decomposed into the tokens “<0xED>”, “<0x96>”, and “<0x84>”, and “버” was decomposed into “<0xEB>”, “<0xB2>”, and “<0×84>” at the byte level. Contrastingly, the tokenizer with an expanded vocabulary tokenized “햄” and “버” in their original forms. The tokenizing results of the existing Llama2 model can lead to the following two problems, as indicated in the Chinese ALPACA~\cite{cui2023efficient}:
\begin{enumerate}
  \item \textbf{Increased token length}: The model cannot represent an OOV using a single token that requires three or four byte tokens. This reduces the possible input length of the model and increases the encoding and decoding times.
  \item \textbf{Duplication of byte tokens}: “햄” and “버” are unrelated tokens; however, they are represented using the same byte token “<0x84>”. Therefore, the model may experience confusion while learning two semantically unrelated words with partially identical representations.
\end{enumerate}
These limitations were overcome by introducing the Korean vocabulary, as shown in Equation~\ref{eq-embedding}. A new embedding was generated by combining the existing Llama2 vocabulary $\mathcal{D}_{L}$ and KoBERT~\footnote{\href{https://github.com/SKTBrain/KoBERT}{https://github.com/SKTBrain/KoBERT}} vocabulary $\mathcal{D}_{K}$.

The KoBERT vocabulary, designed by considering Korean morphemes, consists of $|\mathcal{D}_{K}|=8,002$ words, whereas $|\mathcal{D}_{L}|=32,000$. The union of the two vocabularies has a size of  $|\mathcal{D}|=|\mathcal{D}_{L}\cup \mathcal{D}_{K}|=39,478$. Therefore, the size of the newly added dictionary was $|\mathcal{D}_{R}|=7,478$.

\begin{equation}
    \mathcal{D} = [\mathcal{D}_{L};\mathcal{D}_{R}]
    \label{eq-embedding}
\end{equation}
Here, $\mathcal{D}_{L}$ used the word embeddings trained on the original Llama2, and the newly added word embeddings $\mathcal{D}_{R}$ were randomly initialized.

\subsection{Enriching the knowledge information by MLLM pretraining}
\begin{table*}
\centering
\resizebox{\textwidth}{!}{
\begin{tabular}{ll}
\hlineB{4}
\textbf{Category} &\textbf{LIMA dataset \url{(huggingface.co/datasets/GAIR/lima)}}\\
\hline
\texttt{NE change}    &   \texttt{(EN) I heard north ridge of \textbf{mount Stuart} from my friends, can you tell me more?}\\
                      &   \texttt{(KO) I heard north ridge of `\textbf{Bukhansan Mountain}' from my friends, can you tell me more?} \\ 

\texttt{NE change}    &   \texttt{(EN) How to claim tax back (in USA)?}\\
                      &   \texttt{(KO) How to claim tax back \textbf{in Korea}?} \\
                      \hline
\texttt{topic change} &   \texttt{(EN) What are the primary objections \textbf{Democrats} have to a \textbf{border wall}?} \\
                      &   \texttt{(KO) What is the \textbf{Korean Democratic Party’s} opinion on \textbf{voting rights} for overseas Koreans?}\\

\texttt{topic change} &   \texttt{(EN) How to make \textbf{creepy} food??} \\
                      &   \texttt{(KO) How to make \textbf{bizarre} food??} \\
\hlineB{4}
\end{tabular}}
\caption{Instances of modifications in the English LIMA dataset to reflect the Korean cultural context}
\label{tb-LIMA-examples}
\end{table*}
This section introduces methods to reinforce the word and knowledge information of MLLM via CLM-based pretraining. For queries in the Korean language, the publicly released Llama2 13b model responds in English or alternates between English and Korean (code-switching), indicating a limited Korean expression. However, the content is often accurate when Llama2 responds to a Korean query in English. When asked “이탈리아 수도에 대해 한국어로 소개해줘” (“Introduce me to the Italian capital in Korean”), the model replies, “로마 is the capital city of Italy and...” where the proper nouns “로마” (Rome) and “콜로세움” (Colosseum) are generated in Korean but the detailed explanations are provided in English. This is because the knowledge acquired through pretraining was predominantly in English.

This limitation can be overcome by aligning the knowledge of the Korean and English languages in the MLLM by further pretraining it on a small amount of data. The MLLM expanded with Korean vocabulary was trained on the English and Korean Wikipedia, thereby bridging the extensive English knowledge (accounting for 89.7\%) and limited Korean knowledge (0.06\%). This method aligns with that of the multilingual BERT approach, which was pretrained on the Wikipedia data from 104 languages~\cite{pires-etal-2019-multilingual}.

Equation~\ref{eq-clm} shows that the proposed model was pretrained using CLM. Given an input token sequence $x_{<i}=(x_0,x_1,\ldots,x_{i-1})$ the model predicts the next token $x_{i}$, computes the loss by taking the negative log-likelihood of the predicted token probability, and minimizes this loss.
\begin{equation}
    L_{\!CLM}(\theta)\!=\!
    \mathbb{E}_{x \sim\mathscr D_{\!PT}}\left\{\! -\!\sum_{i}{logP(x_{\!i}|x_{\!<i};\theta,\!\mathcal{D})\!}\! \right\}\label{eq-clm}
\end{equation}
Here, $L_{CLM}(\theta)$ represents the loss function of the language model over the pretraining dataset $\mathscr D_{PT}$, $\theta$ signifies the model parameters, $x_{i}$ is the target token for prediction, and $\mathcal{D}$ refers to the dictionary expanded using the Korean vocabulary. Table~\ref{tb-pretraining-data} lists the specific compositions and sources of $\mathscr D_{PT}$. The loss function accounts for the prediction accuracy of each token within the pretraining dataset. The Korean and English bilingual corpora were adopted in the training method and the model was trained on 33 GB of pretraining data for one epoch with a batch size of 8.

\begin{table}
\centering
\small
\begin{tabular}{lllll}
\hlineB{4}
\textbf{Language} & \textbf{Source} & \textbf{Size(GB)} & \textbf{Content} \\
\Xhline{1pt}
\textbf{Korean}  & \verb|Public|   & \verb|22.41| & news, web\\
                & \verb|WIKI-ko|   & \verb|0.76| & wikipedia\\
\textbf{English} & \verb|WIKI-en|    & \verb|9.92| & wikipedia\\
\Xhline{1pt}
\textbf{Total}        &     & \verb|33.09| \\
\hlineB{4}
\end{tabular}
\caption{\label{tb-pretraining-data} The composition of the pretraining data. The \texttt{Public} data is in (\url{www.aihub.or.kr}) }
\end{table}

\section{Instruction Tuning on LIMA}
The Korean language capability was enhanced by pretraining and the existing knowledge between the English and Korean languages was bridged. However, models trained during pretraining have limited applicability because they are specialized for predicting only the subsequent token. Consequently, high-quality Korean SFT data are required to accurately understand the user intent and generate desired responses. This section describes the method for reconstructing Korean SFT data based on English LIMA and introduces an instruction-tuning approach using these data.

\subsection{Building the Korean LIMA}
The Korean LIMA dataset for SFT was constructed based on a version that underwent machine translation using the English LIMA dataset. Consequently, post-processing was required to address the following issues: (1) discrepancies within the authentic Korean linguistic styles owing to machine translation and (2) exclusion of the Korean cultural context stemming from the characteristics of raw sources in the English LIMA dataset. The Korean LIMA dataset used in this study underwent a human review of the initial machine-translated text and modifications to reflect the Korean cultural context, which involved replacing the named entities and changing the main topic. For the human review process, we recruited the reviewers with Korean as their native language, ensuring that they calibrated all the translated data to the most natural Korean linguistic style.
\\
The raw sources for the English LIMA dataset were posts from the English-speaking community forums, such as Stack Exchange and WikiHow, which reflected the cultural context of English speakers. The cultural context refers to a broad spectrum encompassing everything from daily consciousness to political, economic, and social systems. For instance, a sample from the English LIMA dataset, “How to make banana muffins?” may be irrelevant to the Korean culture because “banana muffins” are neither a popular consumable nor a frequently baked item in Korea. To reflect the Korean cultural context in the dataset, we modified the instances in the English LIMA data that featured Western cultural contexts, particularly the American contexts. These modifications ranged from narrow changes, such as renaming the entities, to broader adjustments, such as entirely altering the dataset topic to fit the Korean context (see examples in Table~\ref{tb-LIMA-examples}).

\subsection{MLLM Training using the Korean LIMA}
The Stanford Alpaca~\cite{taori2023alpaca} is an instruction tuned model based on the Llama trained on 52k instruction data. The corresponding training code is open-source\footnote{\url{github/tatsu-lab/stanford_alpaca}}. We adapted the training script to instruct our model using the Korean LIMA dataset. Instruction-tuning follows the SFT method, wherein prompts are provided as inputs to the model which is subsequently trained to produce the user-desired responses. While this process is similar to pretraining, it differs in that only the output of the prompt is used to compute the loss. This can be mathematically represented as follows:

\begin{equation}
    L_{\!SFT}(\theta)\!=\!\mathbb{E}_{x \sim \mathscr{D}_{\!SFT}}\!\left\{\! -\!\sum_{\!i\in out\!}{logP(x_{i}|x_{\!<i};\theta,\!\mathcal{D})\!}\! \right\}
    \label{eq-sft}
\end{equation}
Where $\theta$ represents the model's parameters, $\mathscr{D}_{SFT}$ denotes the SFT dataset, and $x=(x_{0},x_{1},x_{2}...)$ signifies the token sequence of the template containing the instruction and output. 

\begin{figure}[h!]
\centering
\includegraphics[width=7cm]{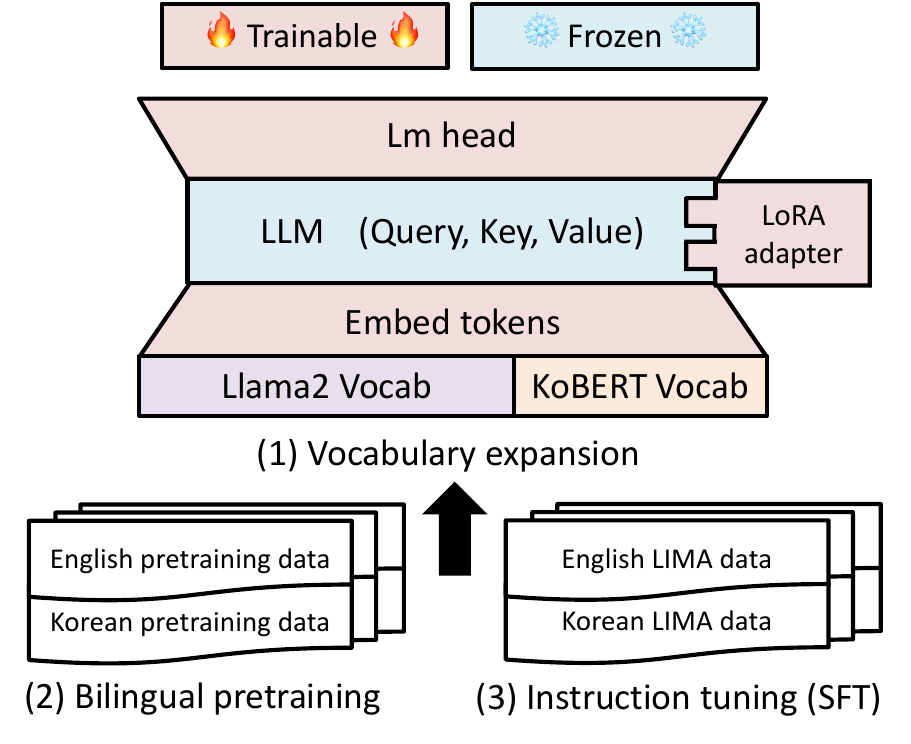}
\caption{\label{fig:training-recipe} Architecture of the proposed model. During training, the LoRA adapter takes Query, Key, and Value from LLM and trains it with new parameters.}
\end{figure}

Pretraining and instruction-tuning require substantial GPU resources. Recent research proposals have focused on training only specific portions of the model that require minimum GPU resources. LoRA~\cite{hu2022lora} is a representative method that involves freezing a pretrained model and infusing each of its layers with trainable rank-decomposition matrices for further training. To apply LoRA, one must choose which parts of the entire model to train. This study trained only the linear layers of the transformer attention, including the query, key, and value, along with the expanded word embedding (as shown in Figure~\ref{fig:training-recipe}).
Consequently, 5.977\% of the total Llama2 parameters were used for training, thereby facilitating the training of our model on a single A6000 GPU.

Figure~\ref{fig:training-recipe} shows the three proposed enhancement methods. The final model underwent the following sequence: (1) vocabulary expansion, (2) bilingual pretraining, and (3) instruction tuning (SFT). Within this context, “Trainable” (red) and “Frozen” (blue) refer to regions where the parameters were updated and not updated during training, respectively. SFT was concurrently performed using the constructed Korean and English LIMA datasets. 

\section{Quantitative evaluation} \label{sec:experiment}
This section describes the following experiments to explore the three research objectives presented in the introduction: (1) comparison between the models with and without the expanded Korean vocabulary; (2) comparison between the model pretrained on the Korean–English bilingual training data and that trained only on the Korean data; and (3) variations in performance owing to instruction-tuning using the LIMA dataset.

\subsection{Evaluation Environment}
\begin{table*}[ht]
\small
\centering\resizebox{\textwidth}{!}{
\centering
\begin{tabular}{lllll}
\hlineB{4}
\textbf{Model} & \textbf{Features} & \textbf{Backbone} & \textbf{Instruction} & \textbf{Pre-training}\\
\hlineB{4}
\href{https://huggingface.co/EleutherAI/polyglot-ko-12.8b}{\texttt{polyglot-ko-12.8b}} & \textbf{Monolingual} model & None & None & 1.2TB  \\
\href{https://huggingface.co/beomi/KoAlpaca-Polyglot-12.8B}{\texttt{KoAlpaca-Polyglot-12.8b}} &+/mono SFT~(21K) &  \texttt{polyglot-ko-12.8b} & 21K & None \\
\href{https://huggingface.co/nlpai-lab/kullm-polyglot-12.8b-v2}{\texttt{kullm-polyglot-12.8b-v2}} &+/mono SFT~(153K) &  \texttt{polyglot-ko-12.8b} & 153K & None \\
\hlineB{4}
\href{https://huggingface.co/meta-Llama/Llama-2-13b-chat-hf}{\texttt{Llama2}} & \textbf{Multilingual} model &  \texttt{Llama-2-13b-hf} & 27K & 2 trillion-token \\
\href{https://huggingface.co/kyujinpy/KO-Platypus2-13B}{\texttt{Ko-Platypus2-13B}} &+/ mono SFT~(25K) & \texttt{Llama-2-13b-hf}  & 25K & None \\
\href{https://huggingface.co/davidkim205/komt-Llama-2-13b-hf}{\texttt{komt-Llama-2-13b-hf}} &+/ mono SFT~(154K) &  \texttt{Llama-2-13b-chat-hf} & 1,543K & None \\
\hline
\texttt{Llama2-koSFT}~(ours) &+/~mono SFT~(1K) &  \texttt{Llama-2-13b-chat-hf} & 1K~(Ko LIMA) & None \\
\texttt{Llama2-ko}~(ours) &+/~mono PT~(33GB) &  \texttt{Llama-2-13b-chat-hf} & None & 33Gb~(Ko) \\
\texttt{Bllossom-ko}~(ours) &\quad+/~expand\_vocab &  \texttt{Llama-2-13b-chat-hf} & None & 33Gb~(Ko) \\
\texttt{Bllossom-bi}~(ours) &+/~bilingual PT, expand\_vocab  & \texttt{Llama-2-13b-chat-hf} & None & 33Gb~(Ko:En=7:3) \\
\texttt{Bllossom-bi-koSFT}~(ours) & \quad+/~mono SFT(1K) & \texttt{Bllossom-bi}(ours) & 1K~(Ko LIMA) & None \\
\href{https://github.com/Anonymous-introduce-paper/LR}{\texttt{Bllossom-bi-biSFT}}~(ours) & \quad+/~bilingual SFT(2K) & \texttt{Bllossom-bi}(ours) & 2K~(Ko-En,LIMA) & None \\
\hlineB{4}
\end{tabular}}
\caption{\label{tb-KoreanLLMs} Overview of the Korean LLMs (The model is from \url{https://huggingface.co})}
\end{table*}

\begin{table}
\centering
\resizebox{\columnwidth}{!}{
\begin{tabular}{@{}l@{}}
\hlineB{4}
\textbf{Prompt} \\
\hline
\smash{질문: 문장 1과 문장 2는 서로 유사한 의미를 가지나요?} \\[-6pt]
\textsubscript{(Question: Do sentence 1 and sentence 2 have similar meanings?)} \\
\smash{문장 1: 습도기 보면 안된다고 경고했어} \\[-6pt]
\textsubscript{(Sentence 1: I warned not to look at the humidifier.)} \\
\smash{문장 2: 습도기 자꾸 보려고 하지마} \\[-6pt]
\textsubscript{(Sentence 2: Don't keep trying to look at the humidifier.)} \\
\smash{정답:} \\[-6pt]
\textsubscript{(Answer:)} \\
\hlineB{4}
\end{tabular}}
\caption{\label{tb-prompt-sts}Evaluation prompt of STS task}
\end{table}

To ensure a fair comparison of LLMs, it is essential to define the task selection for evaluation and specify the LLM model used in the evaluation. To quantitatively evaluate the problem-solving capability of LLMs from various perspectives, the tasks involve language comprehension and inference, sentiment analysis, etcetera~\cite{park2021klue,zhou2023lima}. 
The Korean LLMs were comprehensively evaluated based on eight datasets. 
The benchmarks for the evaluation tasks were KLUE’s NLI, STS, and YNAT, and Naver AI’s Ko-SBI~\cite{lee-etal-2023-kosbi}, and KOBEST's BoolQ, HellaSwag, SentiNeg, COPA~\cite{jang-etal-2022-kobest}, which are described as follows.

\begin{itemize}
    \item{\textbf{Natural Language Inference (NLI)}}: A classification dataset predicting the relationship between two sentences.
    \item{\textbf{Semantic Textual Similarity (STS)}}: A classification dataset measuring the semantic equivalence between two sentences.
    \item{\textbf{YNAT}}: A classification dataset that infers the topic of a given sentence.
    \item{\textbf{SBI}}: A classification dataset aimed at identifying social stereotypes or biases.
    \item{\textbf{BoolQ}}: A question answering dataset for yes/no questions.
    \item{\textbf{HellaSwag}}: A commonsense NLI dataset.
    \item{\textbf{SentiNeg}}: A sentiment classification data.
    \item{\textbf{COPA}}: A classification dataset determining the cause/effect based on a paragraph.
\end{itemize}

The experiments were performed using the Polyglot team’s public branch of EleutherAI’s lm-evaluation harness~\cite{eval-harness} to ensure reproducibility and compare the models. Each model was evaluated using the same data and prompt commands. Table~\ref{tb-prompt-sts} lists the STS evaluation prompts for which each system generated an answer.

For fair evaluation, the model to be evaluated must accurately represent the backbone of the training model and size of the used data, which are defined for the proposed model in Table~\ref{tb-KoreanLLMs}. The Model column is structured in the format ``\texttt{Model}-\texttt{Pretrain Language}-\texttt{Option}''. The \texttt{Pretrain Language} value is `\texttt{bi}' denotes a model that simultaneously uses the Korean and English languages for pretraining. The \texttt{Option} field denotes the application of SFT, where “\texttt{biSFT}” represents the implementation of the Korean and English LIMA data, whereas “\texttt{koSFT}” denotes the usage of only the Korean LIMA data. The \texttt{Bllossom} model refers to the model with vocabulary expansion applied to \texttt{Llama2}. For instance, \texttt{Llama2-ko} is a model pretrained on \texttt{Llama2} in Korean and \texttt{Bllossom-bi} is a model pretrained in Korean and English after vocabulary expansion. The \texttt{Bllossom-bi-koSFT} is a \texttt{Bllossom-bi} model tuned using the Korean LIMA data. \texttt{Polyglot-ko} and \texttt{KoAlpaca}, are presented in Section~\ref{sec:related_work}. 

\begin{table*}[]
\resizebox{\textwidth}{!}{
\begin{tabular}{l|*{10}{c}}
\hlineB{4}
\textbf{Model} &\textbf{NLI}&\textbf{STS}&\textbf{SBI}&\textbf{YNAT}&\textbf{BoolQ}&\textbf{H-Swag}&\textbf{S-Neg}&\textbf{COPA} &\\

&ACC&ACC&F1&ACC&F1&F1&F1&F1&AVG\\ 
\hlineB{4}
\texttt{polyglot-ko-12.8b}               & 35.5 & 50.1 & 48.6 & 31.0 & 59.4 & \textbf{48.8} & \textbf{95.7} & \textbf{81.0} & 56.2 \\
\texttt{KoAlpaca-Polyglot-12.8b}         & 38.0 & 42.7 & 48.4 & 26.0 & 66.4 & 44.4 & 84.8 & 80.0 & 53.8 \\  
\texttt{kullm-polyglot-12.8b-v2}         & 33.9 & 44.8 & 52.5 & 24.6 & 44.2 & 48.3 & 89.8 & 79.3 & 52.1\\
\hlineB{4}
\texttt{Llama2}             & 44.0 & 45.8 & 56.0 & 25.4 & 73.8 & 40.7 & 78.1 & 60.9 & 53.1\\
\texttt{Ko-Platypus2-13B}                & 50.5 & 59.9 & 37.1 & 28.9 & 72.0 & 41.4 & 85.1 & 63.8 & 54.8\\  
\texttt{Komt-Llama-2-13b-hf}             & 33.4 & 51.6 & 48.7 & 24.2 & 52.6 & 39.7 & 62.4 & 64.2 & 47.1\\
\hline 
\texttt{Llama2-koSFT}~(ours)              & 44.5 & 50.6 & 38.5 & 23.1 & 71.7 & 41.2 & 77.3 & 60.5 & 50.9\\
\texttt{Llama2-ko}~(ours)                 & 41.5 & 47.4 & 61.7 & 32.6 & 72.8 & 43.5 & 89.1 & 68.4 & 57.1\\ 
\texttt{Bllossom-ko}~(ours)              & 49.4 & 57.8 & 52.9 & 33.1 & 73.0 & 48.6 & 87.9 & 69.0 & \textbf{58.9}\\  
\texttt{Bllossom-bi}~(ours)              & 48.8 & 46.6 & \textbf{64.5} & 32.8 & 74.0 & 38.0 & 93.2 & 71.2 & 58.6\\
\texttt{Bllossom-bi-koSFT}~(ours)        & \textbf{49.6 }& \textbf{54.9} & 55.0 & 33.9 & \textbf{74.2} & 40.0 & 92.0 & 68.4 & 58.5 \\  
\texttt{Bllossom-bi-biSFT}~(ours)        & 45.7 & 46.4 & 63.4 & \textbf{36.0} & 69.4 & 39.1 & 89.9 & 70.0 & 57.5\\
\hlineB{4}
\end{tabular}
}
\caption{\label{tb-all-results} Benchmarking Korean LLMs: Accuracy (ACC) and F1 score metrics across tasks}
\end{table*}

\subsection{Experiment Results}  \label{subsec:experiment results}
\textbf{(Overall)} Table~\ref{tb-all-results} shows the performances of various models proposed in Table~\ref{tb-KoreanLLMs}. Compared to the monolingual models (such as Polyglot-Ko, KoAlpaca, and Kullm), the proposed multilingual \texttt{Bllossom} models (referred to as “ours”) exhibited an average performance with an increment of approximately 4.57 points. The MLLM performance was affected by the presence or absence of pretraining. The difference between the performances of \texttt{Llama2-ko}, which underwent only pretraining, and \texttt{Llama2-koSFT}, which underwent only SFT, was a substantial 6.2 points. \\

\noindent \textbf{(The influence of vocabulary expansion)}
In Table~\ref{tb-all-results}, \texttt{Bllossom-ko} outperformed \texttt{Llama2-ko} by approximately 1.8 points. For NLI and STS which infer the relationship between two sentences, the \texttt{Bllossom-ko} model with an expanded vocabulary outperformed by 9.15 points. Contrastingly, \texttt{Llama2-ko}, which did not undergo vocabulary expansion, performed better on SBI by 8.8 points. Thus, vocabulary expansion improves the overall comprehension, reasoning, cognition, and causal understanding of the Korean language.
\\ 

\noindent \textbf{(The influence of bilingual pretraining)} 
The \texttt{Bllossom-\textbf{ko}} and \texttt{Bllossom-\textbf{bi}} models in Table~\ref{tb-all-results} differ on the usage of English and Korean bilingual training data during pretraining. The models exhibited similar performances with scores of 58.9 and 58.6, respectively. However, the following observations were made: (1) In contrast to \texttt{Bllosson-bi}, \texttt{Bllossom-ko} exhibited a bias issue wherein the model responded in Korean even when queried in English. (2) For the SBI tasks, \texttt{Bllossom-bi} outperformed by 11.6 points than \texttt{Bllossom-ko}.  And it underperformed 11.2 and 10.6 points on the STS and HellaSwag tasks, respectively. Quantitatively, the impact of bilingual pretraining was minimal; however, a significant performance difference was qualitatively observed owing to bilingual pretraining.\\

\noindent \textbf{(The influence of SFT on the Korean LIMA)} This experiment evaluated the impact of 1K Korean LIMA data by comparing \texttt{Llama2} and \texttt{Llama2-koSFT}, which performed SFT on \texttt{Llama2}.
In Table~\ref{tb-all-results}, \texttt{Llama2}, the backbone, outperformed by an average of 2.2 points. Similarly, the performance of \texttt{Komt}, which underwent an extensive SFT on \texttt{Llama2}, was approximately reduced by six points. The models based on \texttt{Polyglot-ko}, such as \texttt{KoAlpaca} and \texttt{Kullm}, exhibited lower performance than that of the backbone. Therefore, SFT may not significantly influence the quantitative evaluations of classification tasks. However, \texttt{Llama2-koSFT} empirically produced better responses than \texttt{Llama2} based on qualitative factors, such as the quality of the generated responses, vocabulary, and completeness. Therefore, the following section analyzes the effects of SFT based on qualitative evaluations performed by humans~\cite{lee2023qasa} and GPT.

\section{Qualitative evaluation} \label{sec:qulitativeEval}
Based on LIMA~\cite{zhou2023lima}, the qualitative evaluation was performed by humans and GPT. The former involved posing the same question to LLMs \texttt{A} and \texttt{B} and the evaluators subsequently deciding among the responses based on the following three options: Model \texttt{A} is better, Model \texttt{B} is better, or neither is significantly better. Contrastingly, GPT4-based evaluation enabled the GPT to decide among these options. We translated 300 entries from the LIMA human evaluation test dataset into Korean and proceeded with evaluation. According to the LIMA study, the 1k LIMA training data and LIMA human evaluation test dataset were designed to have completely different topics, styles, and tasks. Therefore, tuning the model using LIMA training data negligibly improves the performance~\cite{zhou2023lima}.

\textbf{(Overall)} Figure~\ref{fig:kor_llm_eval_human} and Figure~\ref{fig:kor_llm_eval} show the results of human and GPT4 evaluations, respectively. When comparing \texttt{Bllossom} to \texttt{Koalpaca} and \texttt{Kullm} models of the same size, \texttt{Bllossom} outperformed them in human and GPT4 evaluations and even outperformed the larger \texttt{Llama2-\textbf{70b}-chat} model. Another interesting point is the qualitative evaluation results for human and GPT4 were similar. This was also observed in LIMA.

\subsection{Human-assisted preference evaluation}
\begin{figure}[t]
\centering
\includegraphics[width=\linewidth]{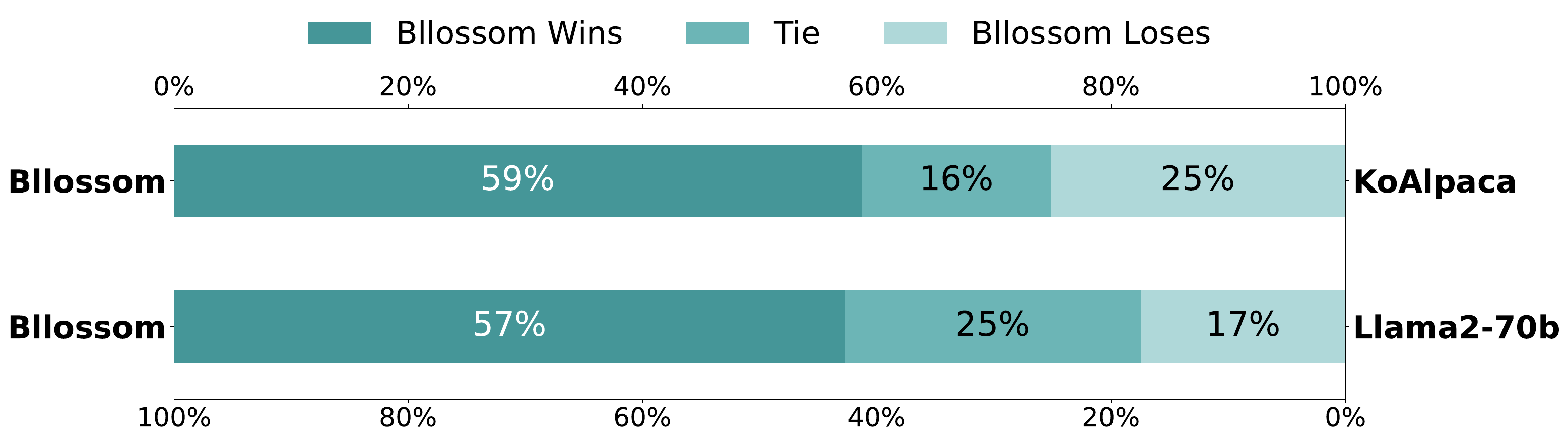}
\caption{Preference evaluation results by human}
\label{fig:kor_llm_eval_human}
\end{figure}

In Figure~\ref{fig:kor_llm_eval_human}, the number of times \texttt{Bllossom} won both \texttt{Koalpaca} and \texttt{Llama2} in human evaluation was 124. The 124 tasks included answering real-world user requests for recommendations, answering questions requiring imagination or creativity, organizing travel itineraries and writing code. In contrast, there were 40 instances where \texttt{Bllossom} gave inferior answers compared to both models, mostly for factual QA. This suggests that \texttt{Bllossom} is not yet as knowledgeable as the larger models, which is likely due to the difference in data size from the pre-training phase.

\subsection{Preference Evaluation using GPT4}
Using the methodology proposed by LIMA, GPT4 was used to compare the performance of \texttt{Bllossom} with six other models. The evaluation was conducted on the Korean LIMA test data.\\

\begin{figure}[t]
\centering
\includegraphics[width=\linewidth]{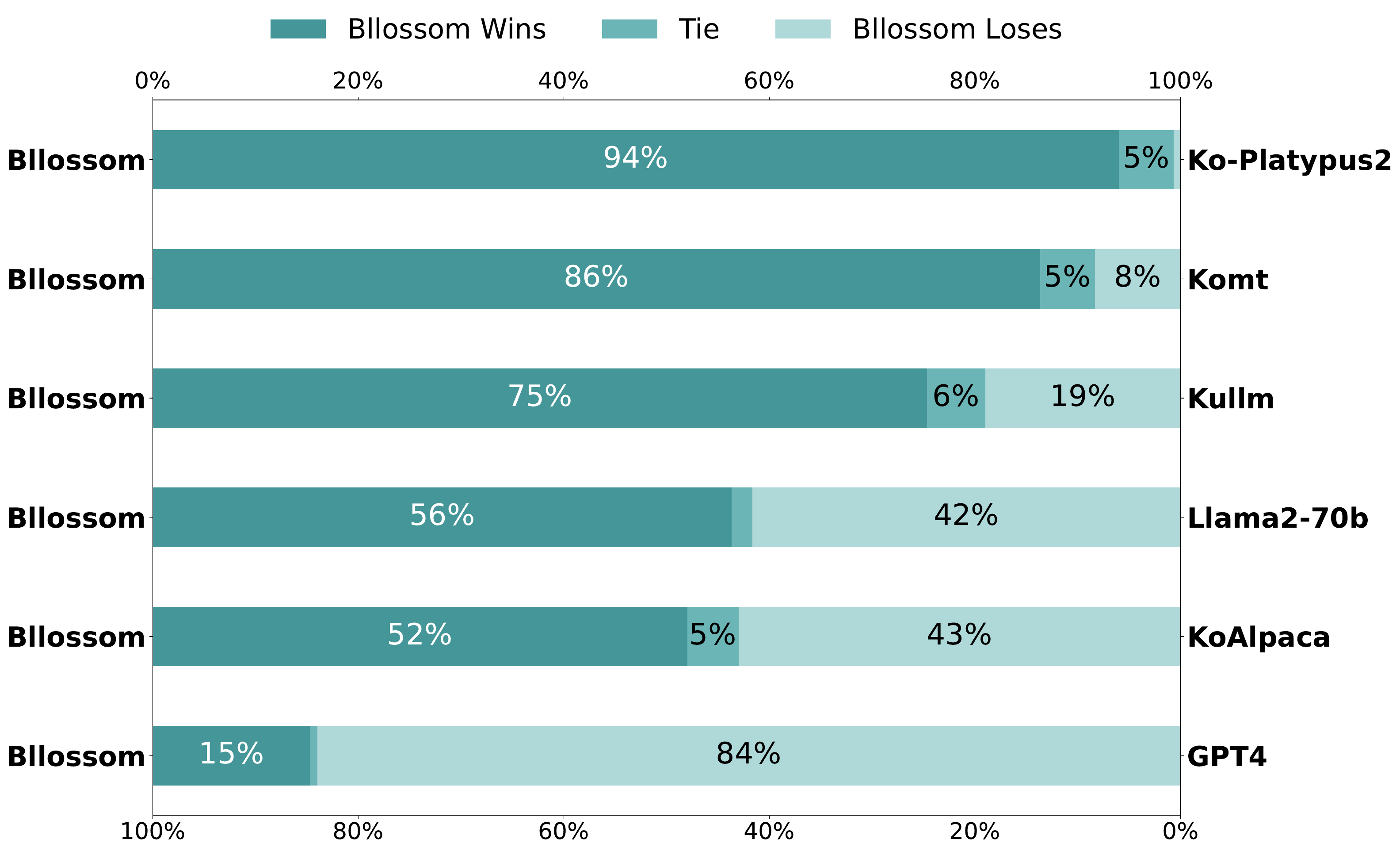}
\caption{Preference evaluation results by GPT4}
\label{fig:kor_llm_eval}
\end{figure}

\noindent \textbf{(Comparing Bllossom with Korean models based on Llama2)} Figure~\ref{fig:kor_llm_eval} shows that \texttt{Komt} and \texttt{Ko-Platypus2} are models based on \texttt{Llama2-13b} that underwent SFT, similar to \texttt{Bllossom}. However, these two models were exclusively subjected to SFT without vocabulary expansion or pretraining. They were fine-tuned using extensive datasets that were either autogenerated or translated, and thus, of lower quality. Qualitatively, \texttt{Bllossom} exhibited superior performance with a margin exceeding 40\%, indicating that pretraining significantly influences the Korean proficiency. During the human evaluation of history-related questions, \texttt{Komt} and \texttt{Ko-Platypus2} either failed to provide answers or exhibited hallucinations more frequently compared to \texttt{Bllossom}. This can be attributed to \texttt{Bllossom} gaining additional knowledge during pretraining.

\noindent \textbf{(Comparing Bllossom with Polyglot-ko-based Korean models)} We discuss whether the proposed \texttt{Bllossom} model exhibits a better qualitative evaluation than Korean monolingual LLMs. \texttt{Polyglot-ko} is a representative Korean monolingual model pretrained on vast Korean datasets and \texttt{KoAlpaca} and \texttt{Kullm} are models trained based on \texttt{Polyglot-ko}. Figure~\ref{fig:kor_llm_eval}, shows that the \texttt{Bllossom} model has a 9\textasciitilde 56\% higher probability of producing superior answers than the two monolingual models utilizing \texttt{Polyglot-Ko} as their backbone. During pretraining, \texttt{Llama2} incorporated a relatively limited set of Korean data; however, its training dataset significantly expanded when English data was included, surpassing the dataset size of \texttt{Polyglot-ko}. This suggests that the bilingual pretraining, which was carried out to augment the deficient proficiency in Korean, has somewhat assisted in bridging the knowledge between Korean and English~\cite{cui2023efficient}.\\

\noindent \textbf{(Comparing Bllossom with GPT4 and Llama)} We discuss the Korean-language proficiency of the proposed \texttt{Bllossom} model. The \texttt{Llama2-\textbf{70b}} model, which has significantly more parameters, was evaluated. Based on the results in Figure~\ref{fig:kor_llm_eval}, the \texttt{Bllossom} model was selected for approximately 14\% of the answers than \texttt{Llama2-\textbf{70B}}. Therefore, in case of an extreme difference in the number of parameters, the differences in performance can be fairly compensated via techniques such as word expansion and pretraining. The qualitative evaluation results for OpenAI’s much larger \texttt{GPT4} model indicated its superiority in frequent answering. \\

\begin{figure}[ht!]
\centering
\includegraphics[width=\linewidth]{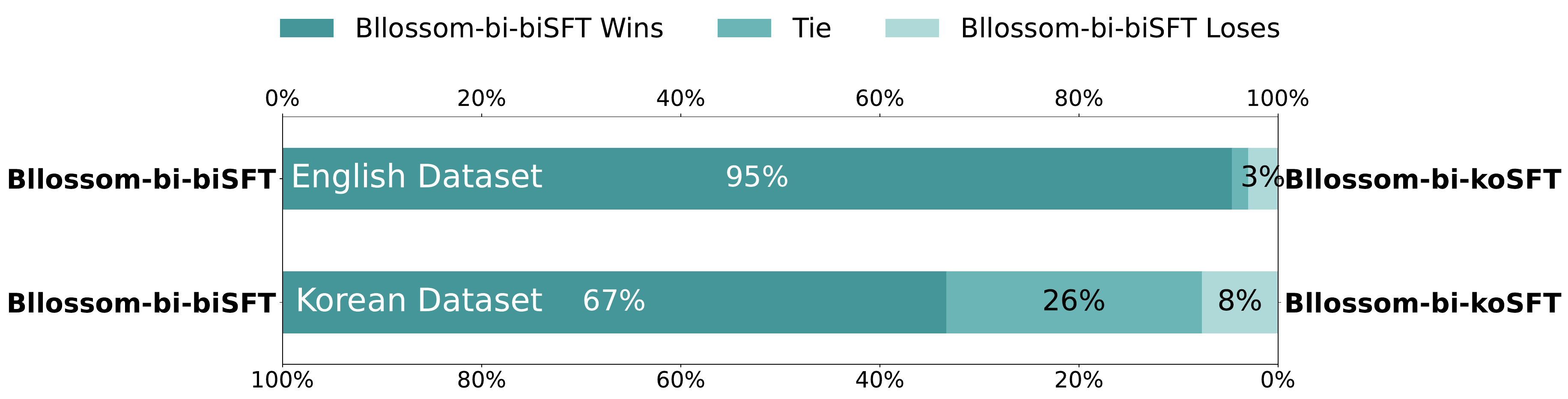}
\caption{Comparing \textbf{bi}SFT and \textbf{ko}SFT models}
\label{fig:instruction_eval}
\end{figure}

\noindent \textbf{(The effect of bilingual dataset for SFT)} In Figure~\ref{fig:instruction_eval}, the \texttt{Bllossom-bi-koSFT} model and the \texttt{Bllossom-bi-biSFT} model differ based on whether bilingual data was utilized for SFT. We conducted a comparative evaluation of the two models using both Korean and English. For the Korean and English LIMA test data, the win ratio for the \texttt{Bllossom-bi-biSFT} model was overwhelmingly high at 67\% and 95\%, respectively. This indicates that contrary to qualitative evaluation, the effect of bilingual SFT in quantitative evaluation is significant. 

\begin{figure}[ht!]
\centering
\includegraphics[width=\linewidth]{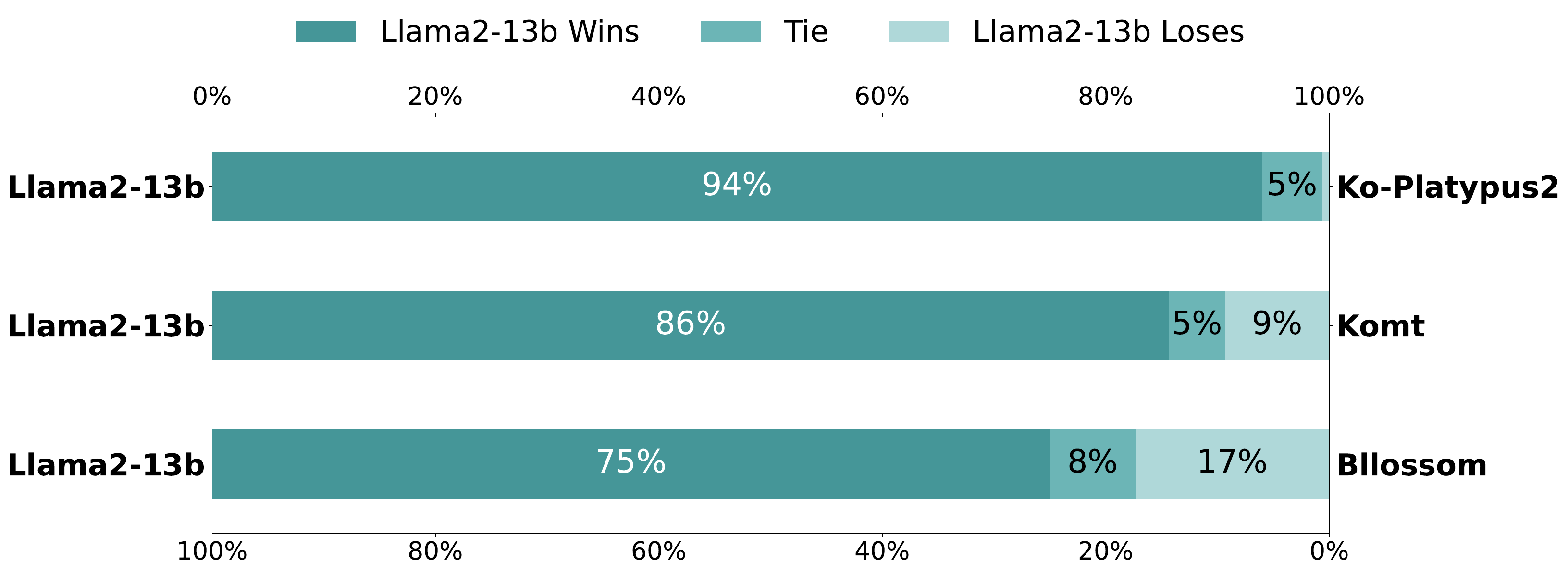}
\caption{Comparing English performance in Llama2-backboned models}
\label{fig:eng_llm_eval}
\end{figure}

\noindent \textbf{(The effect of English language)} 
We investigated whether a model tuned to Korean based on \texttt{Llama2} would perform poorly in English. As shown in Figure~\ref{fig:eng_llm_eval}, all the models appear to have significantly lost their English proficiency compared to the original \texttt{Llama2} model. Nevertheless, \texttt{Bllossom} showed much better performance compared to \texttt{Komt} and \texttt{Ko-Platypus2}. From this, we can infer that while acquiring Korean proficiency, the \texttt{Bllossom} model has a lesser reduction in English capabilities compared to other Korean LLMs.

\section{Conclusion}
This study proposed three methods for enhancing the MLLM capability of LRLs. First, to improve the Korean vocabulary capability of the existing Llama2 model, the vocabularies from KoBERT and Llama2 were merged to create a new embedding. Second, pretraining was performed using bilingual data to enhance knowledge information by aligning high- and low-resource languages. Third, instruction-tuning was performed using the English and refined Korean LIMA datasets to accurately understand the user intent and produce the desired response. Quantitative assessments were performed using eight benchmark datasets and qualitative assessments were conducted using humans and the GPT4 model to investigate the proposed model. The experimental results revealed that the proposed \texttt{Bllossom} model outperformed the pre-existing Korean monolingual models that require vast computing resources and supervised data.

\section{Ethical Considerations}
While we have no ethical concerns regarding the current work, our commitment to upholding the highest ethical standards in all our activities and human evaluations remains unwavering.

\section{Limitations}
In this paper, we proposed a method to enhance Korean language in MLLM. However, to apply the same method to other languages, the following efforts are required. (1) For building LIMA data, one needs to translate 1,030 data, (2) one also need to translate 300 training data for testing.

\nocite{*}
\section{Bibliographical References}\label{sec:reference}

\bibliographystyle{lrec-coling2024-natbib}
\bibliography{lrec-coling2024-example}

\bibliographystylelanguageresource{lrec-coling2024-natbib}
\bibliographylanguageresource{languageresource}

\end{document}